# Predictive analytics for appointment bookings

**MA Nang Laik***

Associate Professor, School of Business

Singapore University of Social Sciences

461 Clementi Road, Singapore 599491

nlma@suss.edu.sg

**Abstract**

One of the service providers in the financial service sector, who provide premium service to the customers, wanted to harness the power of data analytics as data mining can uncover valuable insights for better decision making. Therefore, the author aimed to use predictive analytics to discover crucial factors that will affect the customers' showing up for their appointment and booking the service. The first model predicts whether a customer will show up for the meeting, while the second model indicates whether a customer will book a premium service. Both models produce accurate results with more than a 75% accuracy rate, thus providing a more robust model for implementation than gut feeling and intuition. Finally, this paper offers a framework for resource planning using the predicted demand.

## 1. Introduction

The service company has many years of experience and provides premium service to its customers. Their services are always oversubscribed, and only shortlisted applicants will be notified via mail of the appointment date and time to book the premium service they need. When the customers arrive, they will register themselves and wait for the queue number to be called. The management associate (MA) is also scheduled weekly to meet the customers and offer their professional service.

At the management level, the company wants to know if the customer will show up or not, affecting their human resources planning. If the customers show up, the management wants to see the likelihood that they will sign up for the premium service. The objective is to assign the optimal number of staff to meet all the demands. This goal is even more relevant with Covid19, where the resources are strained and limited.

Based on the current practice, each customer is given X minutes to be served regardless of their chosen service. The manager also assigned Y customers to each MA daily based on their experience. However, there is no existing model to predict if the customers will show up and book the service. Without a reliable predictive model, the company can only assume that customers may show up who may not book the service and waste the company resources. In those cases, customers who are likely to book the service should be given high priority and increases the company's financial status.

Due to the competitive business environment, higher management is keen to make scientific decisions based on data. "Let's data do the talking" is the motto of the management and town hall event for all the employees that have been conducted to convey the messages on the importance of data analytics for future decision making. "Don't come to management meeting without data" is another business norm.

The business objective is to look at past historical data and identify opportunities in the appointment booking to increase the revenue in the current business processes. The research question is to predict the appointment booking status, which has two stages which are sequential in nature. Using the insight that we have gained from the data, we need to predict the customer's appointment status (show verse no show), and when the customers show up for their appointment, will they book the service in the end. The predictive model will create the demand for human resources requirements. Effective planning can lead to better staff satisfaction through balanced workload and efficient utilization of resources while clearing the applicants fast, thereby fulfilling the company objective of providing excellent service to customers and maintaining good customer relationships as a long-term goal.

We have masked sensitive information such as customers' IDs and only reported the finding as generalized information. However, this paper describes all the steps and methods in detail. They will be beneficial to others keen to start on an analytics journey in the organization for process improvements and gain a competitive advantage over its competitors. In section 2, we will do a literature review of the industry and topics related to the predictive model and resource planning. Section 3 will focus on data analysis, and we will build the predictive models for decision making and compute the results. Finally, we indicate the limitation of the model, challenges that we face, and future direction for the research.

**2. Literature Review**

Ma, Choy, and Sen (2014) predicted airline passenger load using various data mining models. It also determined the Decision Tree model as the most useful, with a root mean square error of 3% to 12% observed for all the airlines at the airport. The model was subsequently used to aid in the airport's day-to-day resource planning. On the other hand, Alaeddini, Yang, and Reddy (2011) created a hybrid probabilistic model based on Logistic Regression and Bayesian inference to predict the likelihood of hospital appointment no-shows and cancellations. By using the patients' demographic information and attendance records, the authors could utilize their proposed model to create a selective overbooking system to reduce waiting times. In line with the previous article, Lee, Earnest, Chen, and Krishnan (2005) explored the possibility of Logistic Regression and Decision Trees for predicting hospital failed attendances on a dataset of 22864 patients at Tan Tock Seng Hospital from 2000 to 2004. The predictive model used a cut-off probability of above 0.24 and was able to identify potential defaulters with 80% certainty. Lee et al. (2005) also showed that long waiting periods, repeat defaulters, and younger age groups are associated with the increased likelihood of a 'no-show' situation.

Chandir, Siddiqi, etl. (2018) explored a wide selection of predictive models – Random Forest, Recursive Partitioning, SVM, and C-forest to generate an algorithm that best predicts the children that will not show up for the follow-up immunization visit. The Random Forest model yielded 94.9% sensitivity and 54.9% specificity. Meanwhile, the C-forest, SVM, and Recursive Partitioning models improved prediction with a C-statistic of 0.750 and above. The recursive partitioning algorithm yielded the best result, with an Area under the ROC Curve (AUC) of 0.791.

On the contrary, Gkontzis, Kotsiantis, Panagiotakopoulos, and Verykios (2019) determined the C4.5 algorithm as the best at predicting three different target categories of students' "dropout," "fail," and "no shown up" at final examinations. The C4.5 model had the highest accuracy of 88% compared to the Random Forest, Multilayer Perceptron, and Naive Bayes models. Both Kalka and Weber (2000) recommended induction trees to compute passenger-level no-show probabilities and compared their accuracy with conventional, historical-based methods. Likewise, Hueglin and Vannotti (2001) proposed classification trees and logistic regression to predict both no-shows and cancellations at the passenger level throughout the booking phase at the airline. Finally, Lawrence, Hong, and Cherrier (2003) further explored this topic. They concluded that the Adjusted Probability Model (APM) – a novel method to decipher the output of numerous probabilistic models, was most suitable for predicting airline no-show problems. The new approach displayed the highest accuracy compared to the C4.5 and Naive Bayes algorithms

.

The articles mentioned above have in common the objective of accurately predicting an individual's attendance at an event. Further, all of them concluded the superiority of the predictive models by assessing them in terms of their accuracy rates and AUC. On the contrary, they all proposed the feasibility of different data mining algorithms to organizations for this task. For example, Devasahay et al. (2017), Ma et al. (2014), Lee et al. (2005), Gkontzis et al. (2019), Kalka and Weber (2000) suggested Decision Tree algorithms as the potential classifier to be used to predict a possible no-show. On the other hand, Alaeddini et al. (2011) and Hueglin and Vannotti (2011) narrowed their focus to using Logistic Regression as a potential algorithm to predict a no-show probability. Lastly, Chandir et al. (2018) saw potential in an SVM model, which had outperformed with an AUC (Holdout set) of 0.86, while Lawrence et al. (2003) deemed the APM probabilistic models as the most appropriate for a no-show forecasting problem compared to the Naïve Bayesian algorithm.

Models incorporating specific information on individual customers, such as their demographic profile, can produce more accurate predictions of no-show rates (Lawrence et al., 2003). However, Devasahay et al. (2017) also stated that incorrect data might invalidate the effectiveness of significant variables. Thus, the data preparation step is crucial to ensure the predictive accuracy of the models, which includes removing incorrect data and identifying false entries.

Our contribution in this paper to the research field is three folds. Firstly, we introduce the use of predictive analytics models in the two-stage appointment booking system. Secondly, we identify the important predictors or parameters to predict the appointment show and booked status. Lastly, we developed a holistic approach to link the predictive models to resource planning model. We conducted a study to predict the outcome of appointment bookings for premium service and human resources required to meet the customers' demand. As the business problem is domain-specific and organization-focused, thus required us to build a customized model to address the business problem. Though there are many other studies focusing on predictive analytics and human resources planning, there are limited researches on two-stage predictive model. In addition, the business problem is unique to the organization and therefore requires a different approach to solve. Currently, there is no predictive model available, and everything is performed manually, based on the rule of thumb and gut feeling. In the next section, we explore the data and develop the predictive models.

## 3. Problem Description and Date Understanding

The business objective for this project was to predict the customers' show and book status when they are shortlisted for the appointment for premium service. Therefore, the first business analytics problem would be to build a predictive model to predict whether customers will show up when they are given an appointment. The second business analytics problem would be to create a predictive model to predict whether the customer who shows up will book a service in the end. This report aims to approach these problems by building accurate predictive models for appointment booking to allow for effective human resources allocation at the service counters.

Thus, the data mining goal is to generate the most accurate prediction models that can predict the service booking so that the company can utilize its workforce resources optimally by predicting the customers who may have a higher probability of showing up for an appointment booking service. In this paper, IBM SPSS Modeler will be the primary analytics tool utilized to build the models.

Two datasets, including the past five years of data from 2015 to 2019, were given. There are more than 150,000 customers and 38 variables, most of which are categorical data. In addition, two essential target variables have been derived, such as Show Flag (1/0) and Booked Flag (1/0), based on the historical booking detail.

The following rule has been used to derive the Show and Booked Flag.
If the customers booked the service, show Flag is 1 and Booked Flag is 1
Else if the customers show up but didn't book the service, show Flag is 1 but Booked Flag is 0
Else if the customers didn't show up, then show Flag is 0, and Booked Flag is 0.

There are also cases where customers booked the service but didn't complete the service due to some financial difficulties is considered canceled. But such cases are pretty rare and only take up less than 1% of the data available. We can thus discard these data for the analysis. In the data preparation process, many fields have blank fields. We need to consider replacing them with mode for categorical variable, mean/median for numerical value, or just leaving it blank. There are also outliers in household income, and we have affirmed with the domain expert, which is genuine and valuable in the model; thus, they are not removed.

Table 1: Customers' appointment status

| Quarter | Shortlisted customers | Percentage of 'Show' | Percentage of 'Booked' |
|---|---|---|---|
| 1st half 2015 | 18,011 | 84.70% | 27.20% |
| 2nd half 2015 | 19,716 | 87.80% | 21.00% |
| 1st half 2016 | 20,280 | 87.50% | 20.50% |
| 2nd half 2016 | 17,565 | 87.20% | 23.60% |
| 1st half 2017 | 18,861 | 86.10% | 20.40% |
| 2nd half 2017 | 17,141 | 87.90% | 18.60% |
| 1st half 2018 | 19,059 | 88.70% | 14.50% |
| 2nd half 2018 | 15,113 | 86.60% | 19.90% |
| 1st half 2019 | 16,964 | 88.60% | 16.80% |
| Total | 162,710 | 87.20% | 20.20% |

Figure 1: Shortlisted customers (Show% and booked %)

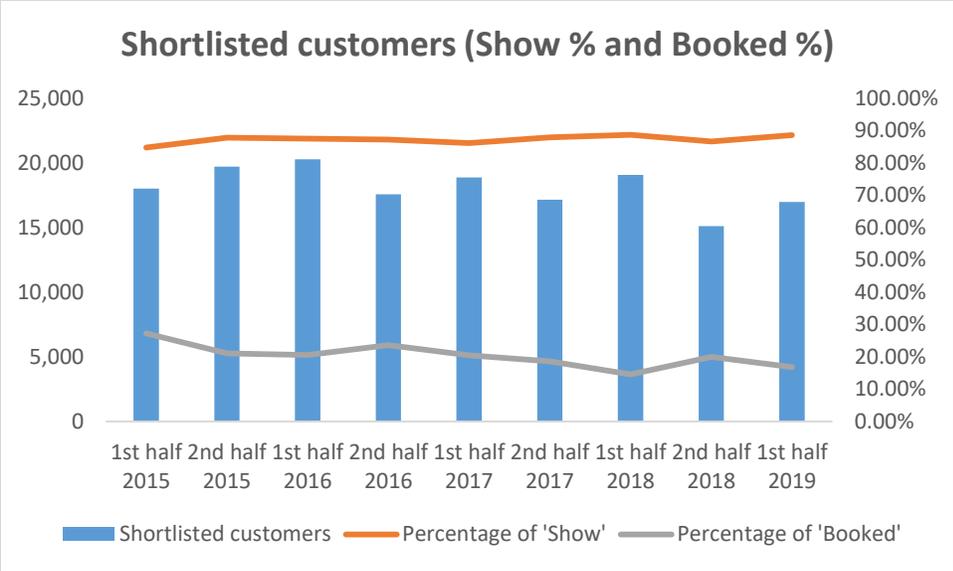

Next, we look at customers' demographic information. 50% of the customers are below 40, 34% are middle-aged, and 16% are elderly above 60. If the company wants to focus on its marketing campaigns, it should focus on the mass 50% even though this group might not be as financially savvy and independent as the middle-aged. But suppose the company can capture the young age group earlier. In that case, the young customer's lifetime with the company is longer. Therefore, it is more beneficial as the existing customers are more likely to buy their services than the new customers. Customers' income information is also analyzed to test the hypothesis where the customers with higher income level is more likely to book the premium service. However, the analysis result showed that there is no difference in income level when customers booked the service. Table 2 and Figure 2 show the percentage contribution (%) by customers' age group.

Table 2: Customer's age group

|  | **Young** | **Middle-Aged** | **Elderly** |
| --- | --- | --- | --- |
| **Age Range** | Below 40 | 40 – 59 | 60 and above |
| **Average Distribution of Age Groups (2015 – 2019)** | 50% | 34% | 16% |

Figure 2: Pie-chart for Customer's age group

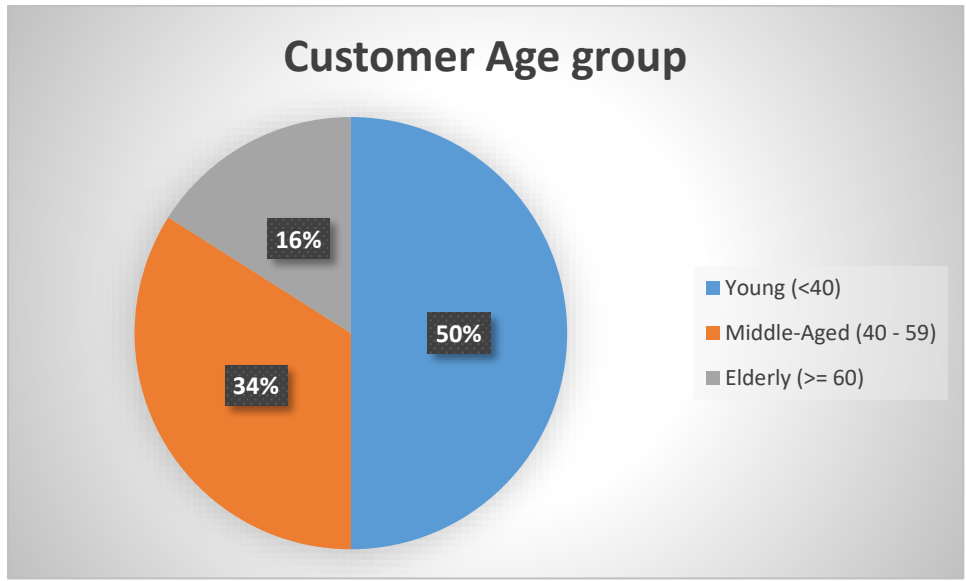

## 4. Predictive modeling

The most commonly used modeling techniques for an event/non-event prediction are the Decision Tree, Logistic Regression, and Neural Network. For example, in the IBM SPSS Modeler, the modeling algorithms for Decision Tree are C5.0, CART, Quest, and CHAID. This paper will use the accuracy rate and AUC as the performance indicators to choose the best model based on the testing dataset.

The first business analytics problem would be to build a predictive model to predict whether shortlisted customers will show up for the appointment. Thus, for the first model, which aims to predict 'Show/No Show', the target variable will be 'SHOW FLAG'. 'BOOKED FLAG' – a variable that is only available after the customers have or have not shown up were excluded for this model. The target variable 'SHOW FLAG' distribution is skewed, whereby the number of 'SHOW FLAG = 0' or 'No Show' cases is only 20% and considered very small compared to 'Show' status. This unbalanced dataset might pose a challenge to the modeling. If we use the dataset as it is to train a predictive model to predict 'SHOW FLAG = 1', it is very likely that one would end up with a model that is almost 100% accurate. The predictive model can predict all cases of 'Show'; while failing to identify all the "No show." It would be a model that does not address the underlying business problem of identifying a potential 'No Show' to optimize human resources. We used the Balance node to increase the proportion of 'No Show' and 'Show' records to a 50:50 ratio. This way, the algorithms will be able to create a model where a fair number of cases are predicted as 'No Show'.

Next, we partition the dataset into a training and testing set. The data was partitioned into 80% for training and 20% for testing as the best models as it had yielded the highest accuracy rates and had the best-performing Area-under-the-curve (AUC) for all models. Other ratios experimented with were: 70%-30% and 90%-10%, respectively. Finally, we used the Auto Classifier node in SPSS to explore every possible combination of options and rank each candidate model based on their AUC and accuracy rates. The top three models identified by the Auto Classifier node were the Neural Net, C&RT, and CHAID– all of which will be further explored and evaluated in the following section. We have also identified the crucial factors to predict the Show and Booked status.

To evaluate the best model, we will use the most common evaluation methods mentioned in the literature review for this paper. Thus, for each target, the champion models will be identified based on their performance in their accuracy rates, hit rates (i.e., the actual occurrence of an event), and their ROC-AUC results. A simple rule-of-thumb for model assessment based on AUC is given as follows: AUC < 0.05 is a poor prediction, AUC between 0.7 and 0.8 is acceptable, AUC of 0.8 to 0.9 is excellent and AUC >= 0.9 is an outstanding model.

An emphasis was given to the models that do well on AUC. It is a general measure of prediction that separates classifier assessment from operating conditions such as misclassification costs (Lessmann & Voß, 2009). Keep in mind that the model with the best AUC may be chosen as the preferred model to evaluate these models. Confusion matrices were also employed to measure the accuracy of the models by comparing the predicted values against the actual target values in the data (SUSS, 2018). Referring to Table 3, comparing the results of Neural Net, C&RT, and CHAID, we can conclude that the **Neural Net** is the **champion model**. It had the best-performing AUC, accuracy, hit, and specificity rates. The Neural Net model also had the highest true positive rate of 68.09%, meaning more 'Show' records were predicted correctly as 'Show' rather than 'No Show'.

Table 3: Table of Comparison for Generated Models for 'SHOW FLAG'

|  | **NEURAL NETWORK** | **C&RT** | **CHAID** |
|---|---|---|---|
| **AUC** | **0.85** | 0.81 | 0.80 |
| **Accuracy Rate (%)** | **78.1%** | 77.3% | 73.5% |
| **Hit Rate (%)** [True Positive Rate] | **68.5%** | 62.7% | 61.7% |
| **Specificity Rate (%)** [True Negative Rate] | **87.6%** | 91.8% | 85.5% |

Since the second business analytics problem would be to build a predictive model to predict if the applicants who show up will book the service in the end, the target variable will be 'BOOKED FLAG'. The customers' data who have shown up for the appointment are selected for this purpose. Those customers' who did not show up for the appointment are discarded as they are unable to book the service. The data was partitioned into 50% for training and 50% for testing as this partition ratio yielded the best accuracy rates and AUC for all models. For 'BOOKED FLAG, the top three models identified by the Auto Classifier node were Neural Net, CHAID, and Logistic Regression (LR) –all of which will be further explored and evaluated.

Table 4: Table of Comparison for Generated Models for 'BOOKED FLAG'

|  | **NEURAL NETWORK** | **CHAID** | **LR** |
|---|---|---|---|
| **AUC** | **0.95** | 0.93 | 0.89 |
| **Accuracy Rate (%)** | 90.3% | 89.3% | 88.6% |
| **Hit Rate (%)** [True Positive Rate] | **92.7%** | 72.8% | 60.5% |
| **Specificity Rate (%)** [True Negative Rate] | 93.2% | 93.5% | 95.7% |

According to Table 4, when we compared the results of Neural Net, CHAID, and Logistic Regression, we can conclude that Neural Net is the champion model. It has the best-performing AUC of 0.951, an accuracy rate of 90.3%, and a high hit rate and specificity rate of 92.63% and 93.17%, respectively.

In this section, we have developed the analytics models and outline how to deal with unbalance dataset. We also share the insights from the data which will help the company to manage the appointment bookings and improve the show up and booking rate. First time buyer with a medium income range are likely to show up and booked the service. The result showed that more than 80% of the customers with this profile will book the service. However, elderly and second time buyer are likely to show up but didn't book the service. With these insights, the company can plan and overbooked elderly second time buyer and give priority booking slot to first time buyer as well. The two stage model approach improve the overall outcome of the appointment booking by more than 20%. With the predictive model, the company can better understand the factors which can affect the customers show up and booked status as explained above. The MA can also take necessary action such as calling or email customers, to remind them of the booking and request them to either confirm or cancel the booking if they no longer required the service. This will reduce the number of no-show and idle time or non-productive slot for the company.

## 5. Framework for human resource planning

Referring to earlier work done by author Ma, Choy (2018), we outline the company's resource planning framework.

We can estimate the demand for the human resource by using the predictive customers' show/no show status and booked/no booking status. We also need to derive the service time required based on the historical data. Given the number of staff, the management can decide how many customers they can serve within the months. If the total time required is more than one month, we can do a what-if analysis to derive the optimal number of staff needed.

Total time available= # of staff * working hours * utilization rate * # of working days per month

Total time required = # of forecasted customer * service time required

Ratio = Total time required / Total time available

If the ratio is more than 100%, the company needs to assign more staff to service the customers. We can also compute the optimal number of staff required given the total demand and time required.

## 6. Conclusion and recommendation

In this paper, we identify the business problem in one of the service company to use data analytics to predictive the show status and booking status of the customers based on two-stage approach. The predictive models aim to capture every aspect of the system, seeking to allow the management to accurately determine the probability of customers showing up and booking a service, which will generate revenue for the company.

The predictive models developed only serve as a guide and will provide actionable insights for the company to make better decision. However, we also assume that it does not show the actual situation as historical data from the past five years is used as an input. However, users can utilize these models developed as a quick and fact-based solution to determine an estimated outcome rather than based on personal' experience or gut-feeling. A prediction system based on Neural networks can be incorporated into the company's human resources system when analyzing the prospect of a customer showing up or booking a service. Further, they can deploy the results by identifying potential customers who are likely to show up or book a service and assign more MA to service the demand. With better planning of human resources, these customers (potentially require more attention and service) will get good service quality and less waiting time. For example, to gauge the customers likely to show up and book a service, the staff can filter out a list of all the customers marked as 'SHOW FLAG = 1' and 'BOOKED FLAG = 1' for the new dataset and assign more experience staff to these customers to improve their chances of booking. The ones most likely to show and book service are offered on the top to narrow the focus to only these valuable customers. However, the models only serve as a decision support tool for the company management. It is up to the management's discretion to make on the ground decisions and manage trade-offs.

In this paper, the results showed the superiority of the Neural Net model in terms of accuracy and predictive effectiveness. When used with its optimal configuration, the Neural Net is a robust method that delivers accurate results due to its ability to model complex non-linear decision boundaries. However, the management would like to implement rule-based decision-making as it is easily understandable and interpretable. Thus, the decision tree model has been selected by users for the deployment. With proper intervention methods, the company is able to improve the financial status of the company using the predictive model and reduce the appointment no-show by twenty percent.

**Acknowledgment**